\documentclass[a4paper,conference]{IEEEtran}
\usepackage{times}
\usepackage{epsfig}
\usepackage{graphicx}
\usepackage{adjustbox}
\usepackage{algpseudocode}
\usepackage{amsmath,amssymb,amsfonts}
\usepackage{textcomp}
\usepackage{xcolor}
\usepackage[ruled,vlined]{algorithm2e}
\usepackage{multirow}
\usepackage{authblk}
\def\BibTeX{{\rm B\kern-.05em{\sc i\kern-.025em b}\kern-.08em
    T\kern-.1667em\lower.7ex\hbox{E}\kern-.125emX}}
\usepackage[backend=bibtex,bibstyle=ieee,citestyle=numeric-comp]{biblatex}
\usepackage{subfig}
\newcommand{\norm}[1]{\left\lVert#1\right\rVert}
\usepackage[font=small,labelfont=bf]{caption}
\DeclareCaptionFont{small}{\small}
\makeatletter
\newcommand{\linebreakand}{%
  \end{@IEEEauthorhalign}
  \hfill\mbox{}\par
  \mbox{}\hfill\begin{@IEEEauthorhalign}
}
\makeatother
\begin{document}
%
% paper title
% Titles are generally capitalized except for words such as a, an, and, as,
% at, but, by, for, in, nor, of, on, or, the, to and up, which are usually
% not capitalized unless they are the first or last word of the title.
% Linebreaks \\ can be used within to get better formatting as desired.
% Do not put math or special symbols in the title.
\title{Adversarial Fine-tune with Dynamically Regulated Adversary}

% author names and affiliations
% use a multiple column layout for up to three different
% affiliations
% \author{\IEEEauthorblockN{Michael Shell}
% \IEEEauthorblockA{School of Electrical and\\Computer Engineering\\
% Georgia Institute of Technology\\
% Atlanta, Georgia 30332--0250\\
% Email: http://www.michaelshell.org/contact.html}
% \and
% \IEEEauthorblockN{Homer Simpson}
% \IEEEauthorblockA{Twentieth Century Fox\\
% Springfield, USA\\
% Email: homer@thesimpsons.com}
% \and
% \IEEEauthorblockN{James Kirk\\ and Montgomery Scott}
% \IEEEauthorblockA{Starfleet Academy\\
% San Francisco, California 96678--2391\\
% Telephone: (800) 555--1212\\
% Fax: (888) 555--1212}}
\author[ ]{Pengyue Hou}
\author[ ]{Ming Zhou}
\author[ ]{Jie Han}
\author[ ]{Petr Musilek}
\author[ ]{Xingyu Li}

\affil[ ]{Department of Electrical and Computer Engineering, University of Alberta}
\affil[ ]{\textit {\{pengyue, mzhou4, jhan8, pmusilek, xingyu\}@ualberta.ca}}
\setcounter{Maxaffil}{0}
\renewcommand\Affilfont{\itshape\small}

% for over three affiliations, or if they all won't fit within the width
% of the page, use this alternative format:
%
% \author{\IEEEauthorblockN{\textbf{Pengyue Hou}}
% \IEEEauthorblockA{
% \textit{University of Alberta}\\
% pengyue@ualberta.ca}
% \and
% \IEEEauthorblockN{\textbf{Ming Zhou}}
% \IEEEauthorblockA{
% \textit{University of Alberta}\\
% mzhou4@ualberta.ca}
% \linebreakand 
% \IEEEauthorblockN{\textbf{Jie Han}}
% \IEEEauthorblockA{
% \textit{University of Alberta}\\
% jhan8@ualberta.ca}
% \and 
% \IEEEauthorblockN{\textbf{Petr Musilek}}
% \IEEEauthorblockA{
% \textit{University of Alberta}\\
% pmusilek@ualberta.ca}
% \and
% \IEEEauthorblockN{\textbf{Xingyu Li}}
% \IEEEauthorblockA{
% \textit{University of Alberta}\\
% xingyu@ualberta.ca}
% % \author{\IEEEauthorblockN{Anonymous Authors}}
% }

% \maketitle

\maketitle

% As a general rule, do not put math, special symbols or citations
% in the abstract
\begin{abstract}
Adversarial training is an effective method to boost model robustness to malicious, adversarial attacks. However, such improvement in model robustness often leads to a significant sacrifice of standard performance on clean images. In many real-world applications such as health diagnosis and autonomous surgical robotics, the standard performance is more valued over model robustness against such extremely malicious attacks. This leads us to the question: to what extent can we improve the robustness of the model without sacrificing standard performance? This work tackles this problem and proposes a simple yet effective transfer learning based adversarial training strategy that disentangles the negative effects of adversarial samples on model's standard performance. In addition, we introduce a training-friendly adversarial attack algorithm, which facilitates the boost of adversarial robustness without introducing significant training complexity. Extensive experiments show that the proposed approach outperforms previous adversarial training algorithms with the following objective: to improve the robustness of the model while preserving model's standard accuracy on clean data.
\end{abstract}

% no keywords

% For peer review papers, you can put extra information on the cover
% page as needed:
% \ifCLASSOPTIONpeerreview
% \begin{center} \bfseries EDICS Category: 3-BBND \end{center}
% \fi
%
% For peerreview papers, this IEEEtran command inserts a page break and
% creates the second title. It will be ignored for other modes.
\IEEEpeerreviewmaketitle

\section{Introduction}
Deep learning has been widely applied to real-world applications with excellent performance, such as autonomous driving, efficient web search and facial recognition. However, recent works have shown that most deep learning models, including state-of-the-art deep neural networks (DNNs), are vulnerable to adversarial attacks~\cite{b1,b2,b3,b4}. These adversarial samples are carefully designed by adding small, human-imperceptible noises to original images, but can lead a well-trained classifier to wrong predictions. To improve model adversarial robustness, many defense strategies, such as data augmentation, gradient masking, adversarial example detection, adversarial training,\textit{ etc.}, have been proposed with the aim of finding countermeasures to protect DNNs~\cite{b4,b5,b6,Madry}. Particularly, adversarial training is widely accepted as the most effective solution. It incorporates adversarial data in model training and helps build model robustness to adversarial attacks. 

Despite its success in improving model robustness to adversarial data, state-of-the-art (SOTA) adversarial training strategies are observed to cause model performance degradation ~\cite{Madry,downgrade1,downgrade2,downgrade3}. For instance, SOTA adversarial training methods such as TRADES~\cite{thero} loses about 10\% standard accuracy for a 50\% adversarial robustness improvement on CIFAR-10 image set. This observation has led to a discussion of the relationship between adversarial robustness and standard generalization (e.g., classification accuracy), with a central debate on whether accuracy and robustness are intrinsically in conflict. Some studies, for example, the works by Tsipras \textit{et al.} \cite{odds} and Raghunathan \textit{et al.} \cite{hurt}, claim that the trade-off of model accuracy and adversarial robustness is unavoidable due to the nature of DNNs. In contrast, Raghunathan \textit{et al.}~\cite{Raghunathan,Stutz} argue that the robustness-accuracy trade-off could disappear with unlimited data. Yang \textit{et al.}~\cite{closerlook} present a theoretical analysis, as well as a proof-of-concept example, showing that this trade-off is not inherent and arguing that the observed accuracy-robustness trade-off is introduced by limitations in current adversary training methods. More recently, Xie \textit{et al.}~\cite{Xie} leverage adversarial samples to improve model accuracy by introducing an auxiliary batch normalization layer particularly designed for adversarial samples. However, this study does not discuss if the model robustness is improved or not. To summarize, although adversarial training improves model robustness against adversarial attacks, how to achieve this goal without trading off model accuracy on clean data in adversarial training is still an open question and remains under explored.

It should be noted that a big loss of standard performance on clean data is unacceptable in many applications that might cause severe consequences. Instead, the standard performance is more valued than model robustness against the extremely malicious adversarial attacks. Such applications include medical diagnosis, autonomous surgical robotics, etc. This leads to the question: To what extent can we boost model robustness without sacrificing standard performance? Unlike prior adversarial training works that allow model performance degradation on clean data, we investigate if it is feasible to improve adversarial performance without any standard performance loss. Specifically, this paper proposes a pre-train based adversarial training strategy, where adversarial training is applied to clean-data (vanilla-trained) models. To prevent model catastrophic forgetting in model refining, we follow the replay-based strategy and maintain the 1:1 clean-adversarial data ratio in model refinement. To further reduce adversarial samples' negative impacts on standard accuracy, we incorporate novel dynamically regulated adversarial (DRA) samples in model adversarial refinement. Unlike most adversarial attacks that add adversarial noise to every image pixel, DRA searches for highly stimulated adversarial features and generates adversarial samples accordingly. Such strategy in adversarial sample generation enforces the model refinement to learn descriptive but non-robust features. Extensive experimentation shows that the proposed adversarial training strategy improves model adversarial robustness with a large margin, but without standard performance loss. %For instance, a ResNet-50 trained with the proposed method achieves 93.77\% standard accuracy, beating its vanilla counterpart by 0.12\% with a considerable increase in robustness against adversarial attacks. 
The contributions of this study are summarized as follows.
\begin{itemize}
    \item We propose a simple, yet generic adversarial training strategy that improves adversarial robustness by a large margin without sacrificing standard accuracy. We argue that the proposed method is significant for applications where standard performance is more valued than adversarial robustness. %We present theoretic analysis and empirical studies on the proposed training strategy, showing that our method helps better at maintaining standard accuracy compare to conventional adversarial training settings~\cite{Madry,downgrade1,downgrade2,downgrade3}.
   % \item Our algorithm is cost-efficient and less time-consuming in practice. For example, the models in our CIFAR-10 experiments can be obtained using one RTX 2080 TI GPU within 6 hours. Since our method can work with transfer learning naturally, it is applicable to many small-sample-size scenarios.  %Different from SOTA adversarial training that requires a long training time and a large amount of clean and adversarial samples, 
    %It only requires a few additional epochs based on a pre-trained model. It is noteworthy that CTAT can be naturally incorporated with transfer learning and thus applicable in many small-sample-size scenarios.
    \item We further introduce an unbounded adversarial attack method, namely DRA. It introduces smaller image distortions and facilitates the adversarial refinement to focus on the learning of most descriptive but non-robust features.
    \item We show that our adversarially trained models exhibit robustness not only to adversarial samples; but also against naturally corrupted images, which suggests its potential for real-world applications.
\end{itemize}

The rest of this paper is organized as follows. Section \ref{sota} presents a brief review of related works. Section III formulates the target problem and specifies our motivations. The technical details of the proposed method are elaborated in Section IV. Section V presents extensive experiments and discussions, then followed by our conclusions in section VI.

\section{Related Works} \label{sota}
\textbf{Adversarial attacks} aim to add small, carefully-designed, human-imperceptible noises to clean data to fool a machine learning model. Depending on the availability of prior knowledge of the target, adversarial attacks can be categorized into two types: white box attacks and black box attacks. Since adversarial samples are transferable, many white box attacks follow the paradigm that exploits a surrogate white box model to generate attack samples ~\cite{b13,b6,PAMI}. This section briefly reviews the first-order white-box adversarial attacks, due to their close relevance to this study. %There are also optimization-based attack models, such as the CW attack \cite{b12}. Due to page limitation, 

%\paragraph{\textbf{White Box Attack.}}
Many white box attacks belong to first-order adversarial methods. Fast Gradient Method (FGM) is an intuitive back-propagation based method to generate adversarial samples~\cite{b2,fgm1} and evolves into a better constraint method, namely Fast Gradient Sign Method (FGSM)~\cite{b3}. FGSM searches for adversarial samples following the direction of the gradients of current parameter values and projects them on images.
\begin{equation}
   x_{adv} = x + \epsilon \cdot sign(\nabla_xL(x, y; \theta)),
    \label{sign}
\end{equation}
where $\epsilon$ is the noise constraint for projecting the attack noises onto $L_\infty$ ball, $x$ and $y$ are the clean input samples and their labels in dataset $D$ and $L(\cdot)$ is the loss function.

Iterative Fast Gradient Sign Method (I-FGSM), also called Basic Iterative Method (BIM)~\cite{b11}, is a multi-step variant of FGSM. It provides stronger adversarial examples but with more computational cost. Projected Gradient Descent (PGD)~\cite{b4} is another iterative variant of FGSM with a random starting point.
\begin{equation}
   x_{adv}^t = \Pi_{S} [ x_{adv}^{t-1} + \epsilon \cdot sign(\nabla_xL(x_{adv}^{t-1}, y; \theta)],
    \label{PGD}
\end{equation}
where $t$ denotes the number of iteration and $\Pi_{S}$ is the clipping function that forces the value to reside in a predetermined range. PGD is widely agreed to be one of the strongest attacks and often used to benchmark model robustness.

%We proposes a new first-order adversarial attack, DRA, in this paper. We show in experiments that compared to PGD, DRA achieves a higher success attack rate but introduces lower noise level image-wise.

%\paragraph{\textbf{Black Box Attack.}}
%Recent works have shown that it is invalid to assume that the model is safe without revealing the parameters of DNNs~\cite{b13,b6}. Transferable adversarial examples~\cite{b13} obtained from attacking ensembled model shows that adversarial samples can transfer among different architectures. They also showed that smaller perturbation adversarial samples tend to have higher transferability. A momentum-enhanced IFGSM method (M-IFGSM)~\cite{b6} further improved adversarial transferability by attacking ensembled models with momentum.

%\subsection{Adversarial Training}
\textbf{Adversarial training} incorporates adversarial samples in model optimization and is considered a very effective method to boost model robustness.
%Among various adversarial defense methods, adversarial training is considered very effective~\cite{b4,b6,Madry}. 
%It harnesses adversarial samples to improves model robustness.
Particularly, Madry \textit{et al.}~\cite{Madry} formulate adversarial training as a MinMax optimization problem,
\begin{equation}
  \mathcal{L}(\theta)= \mathbb{E}_{(x,y)\sim D}[\underset{\delta \in S}{\mathrm{max}} L(x+\delta,y;\theta)],
  \label{minmax}
\end{equation}
Where $\delta$ is the perturbation superimposed on the input, and $\theta$ represents network parameters. The inner maximization tries to find adversarial samples that maximize the loss within the allowed permutation range, while the outer minimization tries to minimize the adversarial loss. %In fact, the goal of adversarial attacks is to solve the inner maximization problem. Meanwhile, the goal of adversarial training is to solve the outer minimization problem. Thus, generation of these worst-case adversarial samples (highest attack success rate) is a critical constraint in adversarial training. 

Since Madry's MinMax optimization is prone to over-fit adversarial samples, many adversarial training methods mix clean data and adversarial data in training.
\begin{equation}
   \mathcal{L}(\theta)= \hspace{0.02in} \mathbb{E}_{(x,y)\sim D} [L(x,y;\theta)
   +\underset{\delta\in S}{\mathrm{max}}L(x+\delta,y;\theta)]. 
   \label{mixAT}
\end{equation} 
Note, in (\ref{mixAT}), adversarial examples is leveraged to regularize the vanilla training on clean data. Recently, Raghunathan \textit{et al.}~\cite{Raghunathan} introduce robust self-training (RST) to balance the vanilla and adversarial loss by a regularization parameter $\beta>0$.
\begin{equation}
   \mathcal{L}(\theta,\beta)= \hspace{0.02in} \mathbb{E}_{(x,y)\sim D} [L(x,y;\theta)
   +\beta\underset{\delta \in S}{\mathrm{max}}L(x+\delta,y;\theta)]. 
   \label{beta}
\end{equation}
Another regularized adversarial training strategy, TRADES~\cite{thero}, is proposed to boost model robustness following the Locally-Lipschitz smoothness constraint.
\begin{eqnarray}
    \label{TRADES}
   \mathcal{L}(\theta,\lambda) &=& \hspace{0.02in} \mathbb{E}_{(x,y)\sim D} [L(f_{\theta}(x),y) \\ \nonumber
    &+& 1/\lambda \cdot \underset{\delta \in S}{\mathrm{max}}L(f_{\theta}(x+\delta),f_{\theta}(x))],
\end{eqnarray}
where $f_{\theta}$ denotes the training model parameterized by $\theta$. Unlike RST computing a adversarial loss between the prediction $f_{\theta}(x+\delta)$ and label $y$ as the regularization term in (\ref{beta}), TRADE regularizes the training by calculating $L(f_{\theta}(x+\delta),f_{\theta}(x))$ from a pair of clean sample $x$ and its adversarial version $x+\delta$. The regularization parameter $\lambda$ determines the trade-off between accuracy and robustness in the overall optimization. It is worth notice that small regularization parameter helps the model emphasize more on accuracy over robustness which closely align to our objective of adversarial training. However, we show later in Section V that even very small value of $\lambda^{-1}$ cannot guarantee models to achieve comparable accuracy to vanilla trained models.

Pre-training is a popular training framework that can help reduce training time or improve accuracy performance for fine-tuning downstream tasks. Jeddi \textit{et al.}~\cite{simple} start with a clean data pre-trained model and fine-tune with PGD adversarial training with the aim of reducing the time cost and overfitting issue~\cite{overfit}. Hendrycks \textit{et al.}~\cite{ft2} adversarially pre-train their model on a downsampled ImageNet and apply adversarial fine-tuning which can significantly improve model robustness on CIFAR datasets compare with adversarial training from scratch. Chen \textit{et al.}~\cite{ft1} show that self-supervised pre-training such as Selfie~\cite{selfie}, Jigsaw~\cite{jigsaw}, also lead to better robustness than traditional adversarial training. In contrast to previous adversarial pre-training strategies, our approach mainly focuses on maintaining accuracy and treats adversarial robustness as an added bonus. More specifically, we incorporate a novel adversary generating method (DRA) to facilitate our goal by reducing the learning complexity of adversarial training.

\begin{table}[t]%\smaller
\caption{Performance of adversarial training methods on MNIST and CIFAR-10. Adversarial accuracy is evaluated with PGD attacks. }
\begin{center}
\begin{tabular}{ c|c c|c c } 
 \hline
   & \multicolumn{2}{c|}{MNIST~($\epsilon=0.3$)} & \multicolumn{2}{c} {CIFAR-10~($\epsilon=8/255$)} \\
% \hline
 %\hline
 %  &\textbf{w/o} attack & \textbf{w/} attack & \textbf{w/o} attack & \textbf{w/} attack  \\
  & $\mathcal{A}_{std}$ & $\mathcal{A}_{rob}$ & $\mathcal{A}_{std}$ & $\mathcal{A}_{rob}$\\
 \hline
 Vanilla & 99.3\% & 0.3\% & 93.0 \% & 0.0\% \\
 \hline
 Madry's~\cite{Madry}  & 99.2\% & 95.6\% & 87.3\% & 47.0\%\\ 
 \hline
 Trades($1/\lambda=1$)~\cite{thero} & 99.3\% & 94.1\% & 86.6\% & 44.3\%\\ 
 \hline
 Trades($1/\lambda=6$)~\cite{thero} & 99.3\% &  96.0\% & 81.2\% & 53.5\%\\
 \hline
 MART~\cite{mart} & 99.1\%  & 96.2\% & 83.4\% & 52.8\% \\

 \hline
\end{tabular}
\label{sotas}
\end{center}
\end{table}

\section{Primitives}
\subsection{Problem Formulation}
In this study, we focus on improving model robustness to imperceptible, in-distribution adversarial samples defined by Fawzi \textit{et al.}~\cite{inDistributionRobustness}. %In-distribution adversarial samples, defined by Fawzi \textit{et al.}~\cite{inDistributionRobustness}, refer to those adversarial data that follow the clean data distributions. 
Briefly, assume that clean data follows a distributing $D$, in-distribution adversarial samples $(x')$ can be also roughly described within $D$. %We argue that focusing on in-distribution adversarial robustness helps model jointly benefit from clean and adversarial samples.

To answer the question: to what extent can we boost model robustness without sacrificing standard performance? We formulate the problem as 
\begin{eqnarray}
\label{newProblem}
 &\mathcal{L}(\theta)= \hspace{0.02in} \mathbb{E}_{(x,y)\sim D}  [\underset{\delta\in S}{\mathrm{max}}L(x+\delta,y;\theta)]& \\
 & \text{s.t.} \hspace{0.1in} \mathbb{E}_{(x,y)\sim D}  [L(x,y;\theta)] \leq \mathbb{E}_{(x,y)\sim D}  [L(x,y;\theta_{std})],& \nonumber
\end{eqnarray} 
where $\theta_{std}=\underset{\theta}{\mathrm{argmin}}\mathbb{E}_{(x,y)\sim D}  [L(x,y;\theta)]$ represents the model with parameter $\theta_{std}$ that yields the minimized standard loss. Comparing the new problem in (\ref{newProblem}) with previous adversarisal training methods in (\ref{minmax}-\ref{TRADES}), the new regularization term in (\ref{newProblem}) explicitly defines the behavior of the model: to improve adversarial robustness without the loss of model's standard performance.

\subsection{Motivation}
To tackle the question in (\ref{newProblem}), we re-examine SOTA adversarial training strategies and obtain an interesting observation.
% Let us start with two interesting observations in the adversarial defense that contradict each other. 
Yang \textit{et al.}~\cite{closerlook} show that many real image sets, such as MNIST, CIFAR-10, SVHN and Restricted ImageNet, are $r$-separated, with the smallest inter-category sample distance being no smaller than 2$r$. Furthermore, their empirical separation distance is 3x-7x larger than the typical adversarial perturbation constraint $\epsilon$ adopted in prior arts, i.e. $\epsilon < r$. In theory, any $r$-separated dataset has more than one classifier that are both accurate and robust up to perturbations of size $r$. %We illustrate this in Fig. \ref{motivation}, where a model associated with the green, solid classification boundary is accurate and robust. On the other hand, on such $r$-separated datasets, SOTA adversarial training methods hurt model accuracy, but this performance trade-off is less notable for complex dataset ~\cite{b4}. For instance, adversarially trained models have 10\% accuracy drop on CIFAR-10 \cite{Madry,closerlook}, whereas only 0.6\% accuracy is traded for 40\% more adversarial robustness on ImageNet \cite{b4}. 
This $r$-separated claim aligns well with adversarial robustness experiments on the MNIST dataset in previous adversarial training studies~\cite{Madry, thero, mart}. However, as summarized in Table~\ref{sotas} where the standard performance and adversarial performance are denoted by $\mathcal{A}_{std}$ and $\mathcal{A}_{rob}$ respectively, % For the simple NMIST dataset, it was not hard to achieve outstanding performance on both adversarial and clean data using SOTA adversarial training algorithms. However, 
on CIFAR-10 which is a more complicated dataset, previous adversarial training often leads to around 10\% standard accuracy drop for models to gain desired adversarial robustness.

%Due to the simplicity of MNIST dataset, we noticed that it was not hard to achieve outstanding performance on both adversarial and clean data. This demonstrates that adversarial training is an effective regularization method that can help models benefit in both accuracy and robustness. However, on the CIFAR-10 dataset, it often leads to around 10\% standard accuracy drop for models to gain desired adversarial robustness as shown in Table~\ref{sotas}. In many real-world applications, the accuracy of DNN classifiers is more valued than their robustness, such as medical diagnosis and autonomous driving. This leads to the question that to what extent can we boost model robustness without sacrificing standard performance?

% We argue that the gap between the two observations, especially the observed accuracy-robustness trade-off, is mainly attributed to the inappropriate harness of adversarial samples in existing adversarial training methods. A common adversarial training strategy is to incorporate adversarial samples in model training from scratch. Although this strategy will achieve the goal in theory, it increases the difficulty in model optimization due to the noisy representation of adversarial perturbations and may cause the optimization process to be stuck in local optima, which eventually hurts model generalization after adversarial training.

%Andrew \textit{et al.} \cite{bugs} first treated adversarial examples as features that derived from non-robust patterns that are highly predictive but incomprehensive to humans. 
From the above observation, we hypothesize that the observed trade-off between adversarial robustness and standard accuracy is due to the limitation of model capacity. The problem of image classification on the MNIST dataset is relatively easy and the models adopted in previous adversarial training studies have enough capacity to jointly benefit from standard and adversarial samples following (\ref{minmax}-\ref{TRADES}). However, for complex problems such as classification on CIFAR-10 dataset, conventional adversarial training in (\ref{minmax}-\ref{TRADES}) increases the difficulty in model optimization due to the noisy representation of adversarial perturbations. In another word, learning both standard features and adversarial features (i.e. those non-robust, yet highly predictive patterns \cite{bugs}) is beyond the capacity of those models (e.g. ResNet-18, ResNet-50, etc.).

Under the hypothesis that model capacity is not enough to simultaneously learn standard and adversarial features, we value clean-data accuracy over adversarial robustness. Thus, we made two modifications to existing adversarial training strategies to tackle the problem formulated in (\ref{newProblem}). Briefly, we propose a transfer learning based adversarial training strategy that starts with a clean data pre-trained model which already has strong generalizability on clean data. To further reduce the learning complexity, we incorporate easy-to-learn adversarial samples in the adversarial fine-tuning. Please refer to the next section for details of the proposed method.
%to reduce the training difficulty and help models benefit from both standard and adversarial features. 
%\begin{itemize}
%    \item We start with a clean data pre-trained model to help the classifier initialize with strong generalizability of clean data.
%    \item We incorporated an easy-to-learn adversary (DRA) in the game of MinMax optimization to reduce the complexity of adversarial features.
%\end{itemize}
%to boost model robustness without trading accuracy. unlike we propose a curriculum transfer adversarial training strategy in this paper. %We hypothesize that early involvement of adversarial samples in training bias model optimization and cause the optimization process to be stuck in local optimums on the non-convex loss surface, which eventually hurt model generalization after adversarial training. 

\begin{figure}[t]%\smaller
\centerline{\includegraphics[width=.9\linewidth]{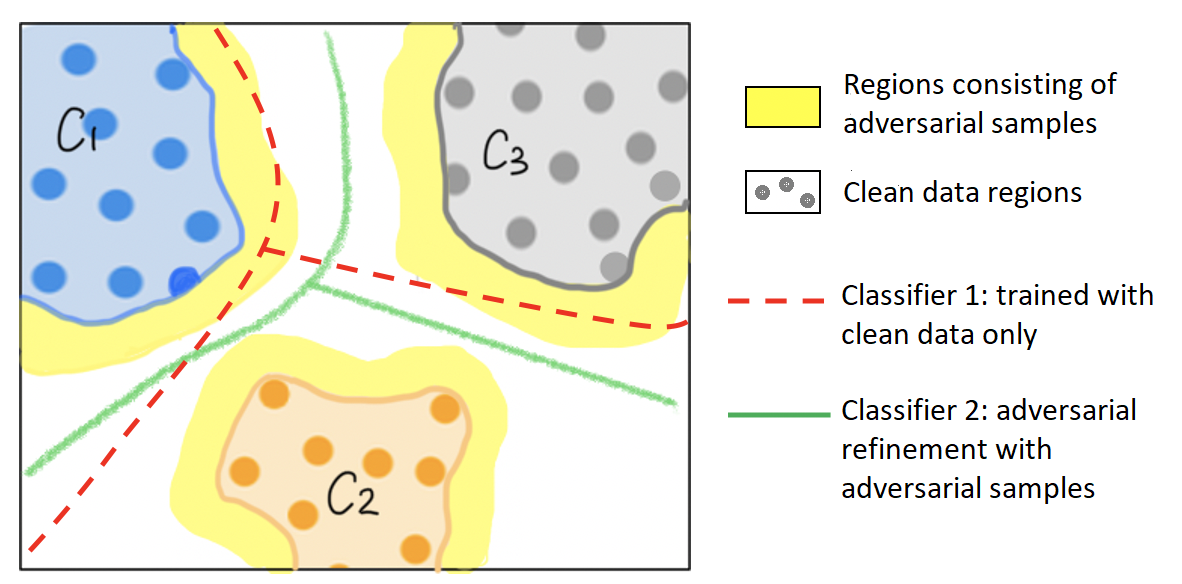}}
\caption{A conceptual demonstration of our adversarial fine-tune method. Classifier 1 represents the clean data pre-trained classifier which is accurate but not robust to adversarial samples. Our adversarial fine-tune method seeking for both accurate and robust classifier by pushing the classifier 2 out of the yellow adversarial regions defined by $\delta \in S$.} % adjusting a pre-trained network to its optimal state (classifier 2).}
\label{motivation}
\end{figure}

\captionsetup[figure]{font=small}
\begin{figure*}\smaller
\centerline{\includegraphics[width=0.9\linewidth]{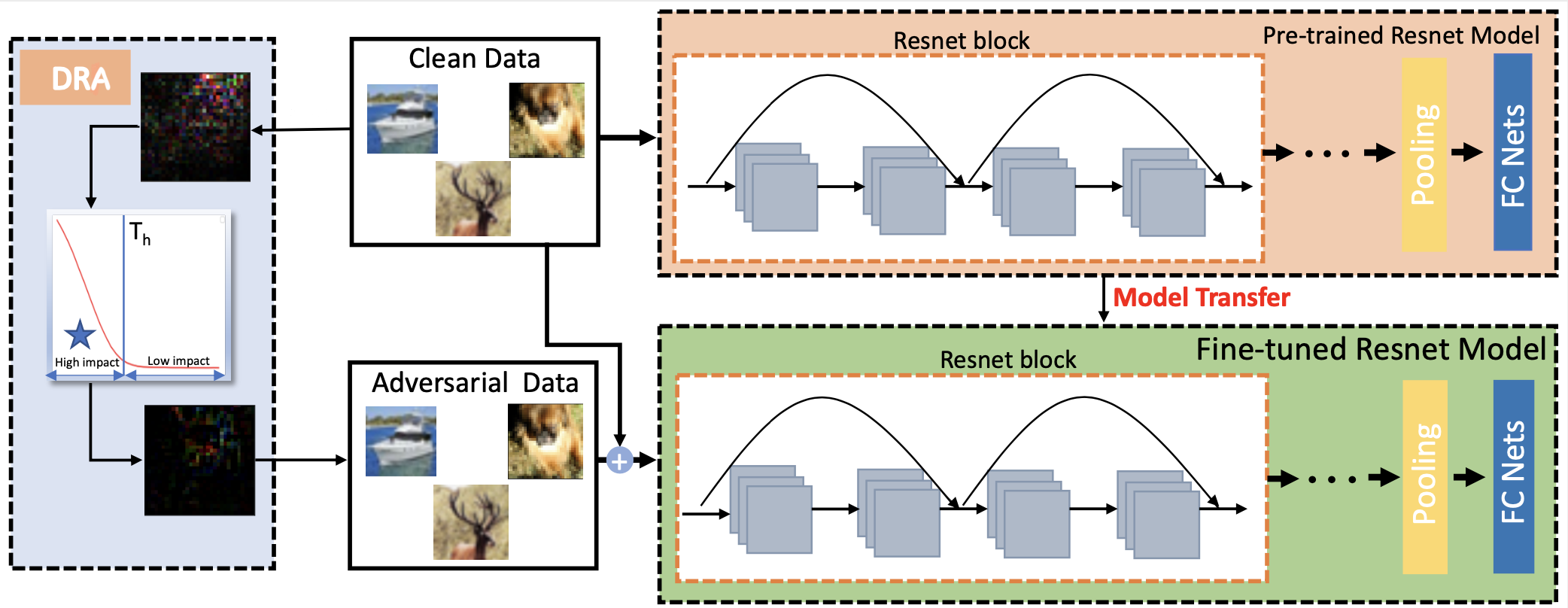}}
\caption{Systematic diagram of the proposed adversarial transfer adversarial training strategy, where we use various ResNets as the backbone. From left to right: DRA, a proposed method to generate adversarial samples. It filters out negligible adversarial noises and reduces adversarial training complexity. The orange block and green block represent standard training and robust training, respectively. in standard training, we train a model with $\theta_{std}$ on clean data that yields high standard performance. Then the robust training aims to find a better $\theta$ in the vicinity of $\theta_{std}$ to boost adversarial robustness without standard performance loss.} %In standard training, we aim to initialize the model with prior knowledge regarding the distribution of original data, i.e. Classifier 1 in Fig.~\ref{motivation}. Then the clean-data pre-trained model is further refined using a more complex image set consisting of clean data and DRA-generated adversarial samples. Later we show that it can help mitigate the negative effects of forgetting pre-trained knowledge while optimizing for model robustness.} %The key idea of our method is to fine-tune the source model using both clean and adversarial data; therefore ensuring the model's generality over clean data and improve robustness at the same time.}
\label{defense}
\end{figure*}

\section{Methodology}
%In this section, we provide a specific description and theoretical analysis of the proposed adversarial fine-tune based training strategy. Particularly, we focus on improving model robustness to imperceptible, in-distribution adversarial samples defined by Fawzi \textit{et al.}~\cite{inDistributionRobustness}. %In-distribution adversarial samples, defined by Fawzi \textit{et al.}~\cite{inDistributionRobustness}, refer to those adversarial data that follow the clean data distributions. 
%Formally, assume that clean data follows a distributing $D$, in-distribution adversarial samples $(x')$ can be also roughly described within $D$. We argue that focusing on in-distribution adversarial robustness helps model jointly benefit from clean and adversarial features.

% \captionsetup[figure]{font=small}

%for $r$-separated datasets, where $r > \epsilon$. In-distribution adversarial samples, first defined by Fawzi \textit{et al.}~\cite{inDistributionRobustness}, are those adversarial data that follow the clean data distribution of the same labels. Formally, assume that clean data and their labels $(x,y)$ follows a distributing $D$, in-distribution adversarial samples $(x+\epsilon,y)$ can be also described by the distribution $D$. We argue that focusing on in-distribution adversarial robustness makes sense, as adversarial perturbations in practice are typically imperceptible to humans. 

\subsection{Transfer Adversarial Training}
SOTA adversarial training methods usually train a robust model from scratch using either adversarial data only in (\ref{minmax}) or a combination of clean and adversarial data following (\ref{mixAT}-\ref{TRADES}). However, none of them guarantee that the standard performance will be preserved. Indeed, due to the highly non-convex loss surface in model optimization, optimizing both targets simultaneously may be at odds with each other \cite{odds}.
%Madry's adversarial training framework adopts only adversarial data for robustness optimization (Eqn.~\ref{minmax}). Due to the highly non-convex loss surface in model optimization, optimizing both targets simultaneously may be at odds with each other \cite{odds}. To overcome this challenge, some works adopted the idea of adversarial fine-tuning and achieved higher adversarial robustness than training from scratch \cite{simple, ft1, ft2}. Compared to previous works, our approach focuses on a different perspective, i.e., pre-training with clean data to maintain accuracy of the model.

To solve the primal conditioned optimization problem in (\ref{newProblem}), we apply the Karush-Kuhn-Tucker (KKT) approach and obtain a dual unconstrained optimization problem by introducing a KKT coefficient $\lambda$,
\begin{eqnarray}
 \mathcal{L}(\theta,\lambda)&=& \hspace{0.02in} \mathbb{E}_{(x,y)\sim D}  [\underset{\delta\in S}{\mathrm{max}}L(x+\delta,y;\theta)] \\
 & +& \lambda\mathbb{E}_{(x,y)\sim D}[L(x,y;\theta) -L(x,y;\theta_{std})]]. \nonumber
   \label{newProblemKKT}
\end{eqnarray}
Therefore, the solution to the dual problem
\begin{eqnarray} 
\underset{\theta}{\mathrm{min}} \underset{\lambda,\lambda>0}{\mathrm{max}}\mathcal{L}(\theta,\lambda)
\label{KKT}
\end{eqnarray}
is identical to the solution of the primal problem in (\ref{newProblem}). Note that to solve the problem in (\ref{KKT}), we need the margin value $\mathbb{E}_{(x,y)\sim D}  [L(x,y;\theta_{std})]$ which is fixed before the adversarial training described in (\ref{newProblemKKT}). That is, a model with $\theta_{std}$ is already obtained for high standard performance. Therefore, instead of training a robust model with $\theta$ from scratch, we introduce a transfer learning strategy to solve (\ref{newProblemKKT}) and propose to search a new $\theta$ from $\theta_{std}$, as the standard performance is unlikely to change severely in the small vicinity of $\theta_{std}$. Fig. \ref{motivation} visualize the conceptual idea of the propose adversarial training strategy on a $r$-separated dataset. Classifier 1 is vanilla-trained over clean data. It is accurate but nor adversarial robustness. By pushing the classification boundary out of the yellow adversarial regions defined by $\delta \in S$, the obtained classifier 2 has the identical standard performance with improved adversarial robustness.

We depict the detailed systematic diagram of the proposed adversarial training strategy in Fig. \ref{defense}. Specifically, the proposed training strategy divides the training into two phases: vanilla standard training and adversarial robust training. In standard training, we exploits clean data to train an accurate model, 
$\theta_{std}=\underset{\theta}{\mathrm{argmin}}\mathbb{E}_{(x,y)\sim D}  [L(x,y;\theta)]$.
%Where $x$ and $y$ are clean input samples and their corresponding labels, $\theta$ is the network parameters. This step helps our model initialized with high classification accuracy but still vulnerable to malicious attacks.
The standard training has two benefits. First, it provides the margin value $\mathbb{E}_{(x,y)\sim D}  [L(x,y;\theta_{std})]$ in (\ref{newProblemKKT}). Second, due to inherent transfer learning property, the downstream adversarial robust training is more cost-efficient than SOTA adversarial training strategies that often require large training loads and long training time to handle the complexity introduced by adversarial features. 

With the clean-data pre-trained model, adversarial samples are incorporated in the model refinement phase. To prevent model catastrophic forgetting in model fine-tuning, we follow the replay-based strategy and let the network iteratively update upon clean and adversarial samples. 
%We depict the systematic diagram of the proposed strategy in Fig.~\ref{defense}, where the orange block and green block represent standard training and robust training, respectively. One may notice that the proposed method aligns with the paradigm of transfer learning, where the standard training and the robustness training can be considered as the source and target tasks, respectively. 
Note, unlike conventional fine-tuning tends to unfreeze several outer layers to preserve the knowledge learned from the source task, we argue that it is important to update all parameters in robust training stage. Briefly, adversarial noises propagate through each layer in the model and aggregate into large prediction distortions. All layers of a robust model must contribute to the defend against adversarial attacks.

\subsection{Dynamically Regulated Adversary in Adversarial Training}\label{DRA}
Adversarial training is usually referred to as the "MinMax" optimization game, where the adversarial samples are significant contributors. In fact, aggressive adversarial attacks are preferred in adversarial training, because models trained on aggressive adversarial attacks are more resistant to weaker adversaries. Therefore previous studies have usually adopted the PGD attack in adversarial training. 

A good adversarial attack approach for adversarial training should be aggressive but without greatly increasing training complexity. Although PGD is an aggressive solution for introducing noise, we argue that PDG is not the best candidate for adversarial training in the sense that it introduces excessive noisy information in model robustness training. More specifically, PGD uses a Sign() clipping method to project adversarial noises onto the $L_\infty$ ball. It treats all image pixels equally and applies the same noise injection strategy to all pixels, regardless of their contribution to the prediction results. We argue that the treatment of adding adversarial noise to all pixels in PGD significantly increases the training complexity. We will show in the experimentation that the proposed DRA attack is a better candidate than PGD in adversarial training and facilitates various adversarial training.  % in order to train models with higher robustness, aggressive adversarial attacks are prefered in adversarial training.%  the inner max function suggests that adversarial samples should try to achieve the highest attack success rate with respect to the current model parameters. 
% Adversarial samples are significant contributors in model robust training. A model trained on aggressive adversarial data is more resistant to weaker adversaries. We argue that a good adversarial attack method in adversarial training should be aggressive but without greatly increasing training complexity.

%PGD is a common method to generate adversarial data in adversarial training. It uses a Sign() clipping method to project adversarial noises onto the $L_\infty$ ball. 
%However, we argue that a good adversarial attack method for adversarial training should be aggressive but without greatly increasing training complexity. Although this is an aggressive solution for introducing noise within the $L_\infty$ bound, we argue that these noisy representations in PGD-generated adversarial data complicate model training. %by introducing excessive information beyond the capabilities of most DNNs. 
%More specifically, PGD applies the same noise injection strategy to all image pixels regardless of their contribution towards prediction results. From this perspective, we believed that PGD is not an ideal adversary generation method to accommodate our objective.

This work introduces a novel gradient-based attack method, namely DRA attack, in Algorithm~\ref{DRA}. Unlike PGD, which treats image pixels equally, DRA distinguishes important pixels from others by aggressively modifying only those image features that are highly predictive, but non-robust. In this way, DRA adversaries enforce the robustness training by focusing more on these predictive features, thus helping to improve model's robustness against adversarial attacks. Specifically, DRA quantifies pixel significance by the gradient of the loss function with respect to its pixel value. A large gradient value suggests that the pixel contributes more to the image prediction. In this regard, DRA abandons the $Sign(\cdot)$ method so that it can smoothly search for the optimal adversarial samples along the gradient. In addition, when generating adversarial samples, DRA uses a more resilient distance metric $L_1$ to bound the adversarial noise instead of using the $L_{\infty}$ constraint.  

\begin{figure}%\small
\centerline{\includegraphics[width=.9\linewidth]{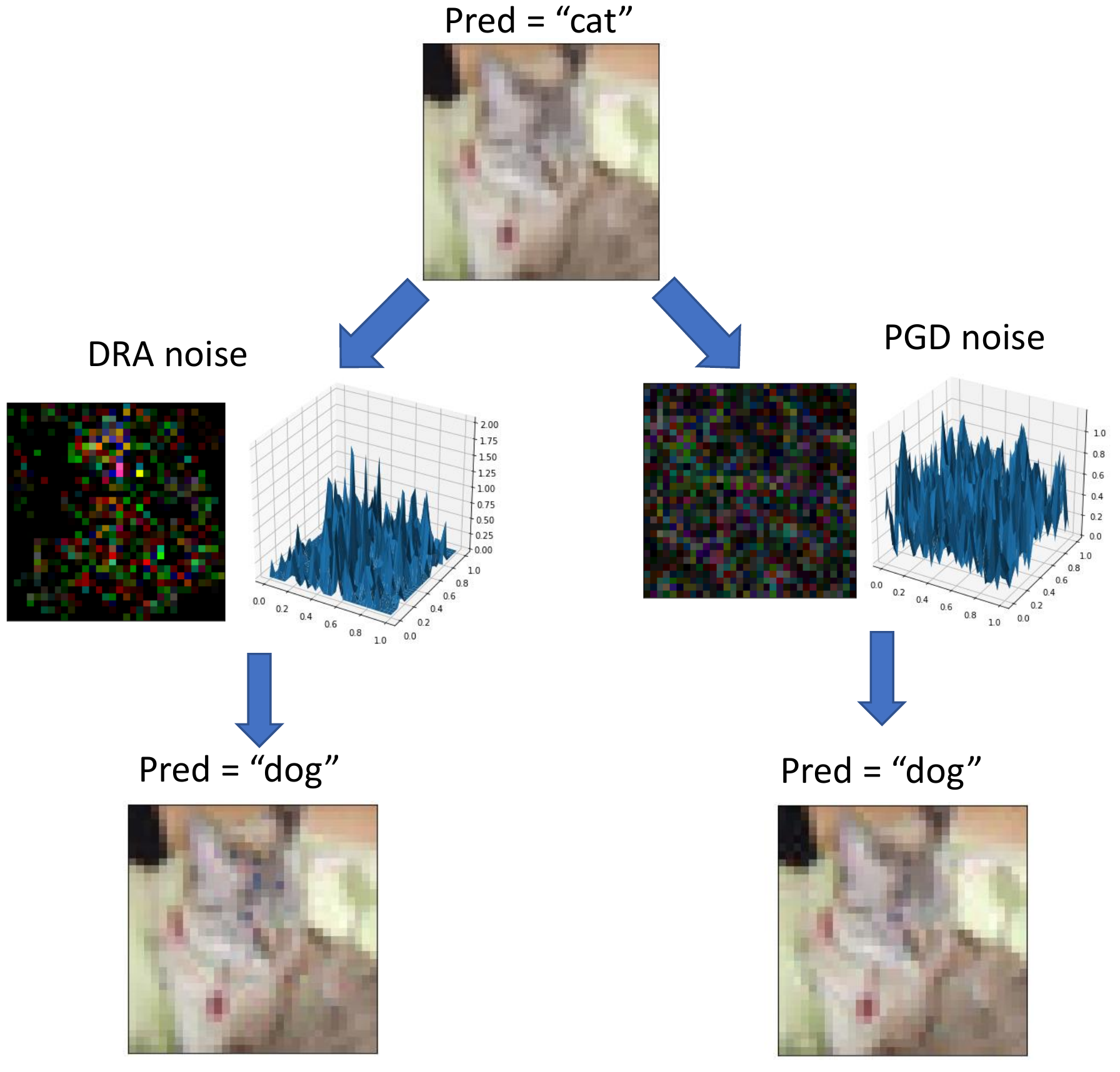}}
\caption{A visual comparison of PGD (left) and DRA (right) adversaries. They are both able to fool DNNs with imperceptible noises, however, the overall noise budget of DRA is smaller. Furthermore, DRA focuses on highly discriminate pixels (patterns that contribute most to final prediction outcomes), where PGD equally distributes adversarial noises over the whole image. As we can see from the example, DRA noises align well with the salient map of the cat.}
\label{cat}
\end{figure}

Fig.~\ref{cat} presents a visual comparison of the DRA and PGD adversaries. Compared to PGD, DRA introduces stronger noise in the "cat" region, which is the most predictive pattern in the image. We claim that the lower noise level in the image background makes the learning more complex. We show in the experimental section that DRA samples help various training strategies achieve higher adversarial robustness than PDG for the same adversarial aggressiveness.

\begin{algorithm}
\SetAlgoLined
\SetKwInOut{Input}{Input}
\SetKwInOut{Output}{Output}
\SetKwFor{For}{for (}{) $\lbrace$}{$\rbrace$}
\Input{Clean-data pre-trained model $f$, loss function $L$, clean training pairs $(x,y)$, regulation term $\epsilon$, significant pixel percentage $p$, iteration $T$}
\Output{Adversarial sample $x^*$}
 1: $\alpha = \epsilon/T$\;
 2: $x_0^* = x$, $t = 0$\;
 3: \If{$t < T$}{
    Input $x^*_t$ to $f$ and obtain gradient $\nabla_x L(x_t^*,y)$\;
    update $x_{t+1}^*$ = $x_t^* + \frac{\alpha \cdot \nabla_x L(x_t^*,y)}{\norm{\nabla_x L(x_t^*,y)}_1}$
}
4: Rank current pixel values based on $\nabla_x L(x_t^*,y)$ and set threshold $T_h$ for the $p$ most significant values\;
5: \If{$x_{ij}^*<T_h$}{
$x_{ij}^* = x_{ij}$\;
}
\caption{DRA attack}
\label{DRA}
\end{algorithm}

%Unlike PGD which adds adversarial noise at every pixel, DRA filters out insignificant pixels and superposes adversarial noise on significant features only. We show in Section~\ref{sec:exp} that although the proposed DRA achieves higher attack success rate, it introduces fewer noises in adversarial samples than PGD. Consequently, DRA-generated samples bring in much smaller negative effects on model generalization in adversarial training.

\section{Experiments}
\label{sec:exp}
In this section, we present extensive experiments to evaluate the accuracy-robustness performance of the proposed adversarial training strategy on MNIST and CIFAR-10 datasets.

\subsection{Experimental Setup}
In the proposed adversarial training strategy, we first train a model on clean data only so that the model has high standard performance. The model is then fine-tuned with clean and DRA samples. In both training processes, we utilize image augmentation including random cropping and random flipping. In addition, the image batch size is set to 128, SGD with momentum to 0.9 and weight decay to $2e-4$ for model optimization. 
%Before adversarial fine-tuning, the model already has good accuracy performance by pre-training on a clean dataset. 

On the MNIST dataset, we follow the TRADES study and use a simple CNN model with 2 convolutional layers and 2 fully connected layers. We set $\epsilon=0.1$, $p=2/3$ and iterate 20 times to generate DRA samples for 50 epochs of robust fine-tuning. 

On the CIFAR-10 dataset, we use ResNet as the backbone model and perform 60 epochs of adversarial fine-tune in the robust training phase. Adversarial samples used for our robust training are generated in 5 iterations. The noise constraint is linearly decayed from 2 to 0.5 in the first 50 epochs, and only clean images are fine-tuned in the last 10 epochs.

For evaluation purposes, adversarial samples are generated by PGD under various noise constraints (e.g. $\epsilon=0.1, 0.3$ out of 1 for MNIST images and $\epsilon=2,5,8$ out of 255 for CIFAR-10) in 20 iterations with step size $\epsilon/20$, unless otherwise specified. We compare our DRA fine-tuned model with TRADES~\cite{thero} on both datasets using relatively small regularization parameter ($1/\lambda$). We find that this setting is consistent with our objective, as the small regularization parameter intuitively enforces TRADES to primarily optimize the clean data accuracy while "trading off" a small amount of adversarial robustness.

%\subsection{Adversarial Defense Evaluation}\label{SCM}

\subsection{Accuracy-Robustness Performance} 
The numerical results on the MNIST are shown in Table~\ref{MNIST exp1}, where "natural" indicates the standard performance of the model on clean data and "vanilla trained" indicates that the model is trained with clean data only. On the MNIST dataset, both TRADES and our method maintain high standard accuracy while improving model's adversarial robustness; the proposed method obtains 2\%-5\% higher robustness than TRADES with various settings. 

\begin{table}[t]%\smaller
\caption{Accuracy-Robustness performance against PGD attacks on MNIST. Note, the target is to improve adversarial robustness without sacrificing standard performance.}
%\begin{adjustbox}{width=0.7\columnwidth,center}
\centering
\begin{tabular}{ c|c|c|c } 
 \hline
  Model & natural & PGD$_{\epsilon=0.1}$ & PGD$_{\epsilon=0.3}$ \\
 \hline
 \hline
 Vanilla trained  & 99.3\%  & 88.3\% & 18.6\% \\
Trades($1/\lambda=1$) & 99.3\%  & 98.4\% & 93.5\% \\
Trades($1/\lambda=0.5$) & 99.3\%  & 97.9\% & 92.1\% \\
Trades($1/\lambda=0.1$) & 99.4\% & 97.2\% & 90.8\% \\ 
%Madry's & 99.1\% & 98.4\% & 96.1\%  \\
Ours & 99.3\% & 98.0\% & 95.9\%\\
 \hline
\end{tabular}
\label{MNIST exp1}
%\end{adjustbox}
\end{table}

\begin{table}[t]%\smaller
\caption{Accuracy-Robustness performance against PGD attacks on CIFAR-10, with ResNet-50 as the backbone model.}
%\begin{adjustbox}{width=\columnwidth,center}
\centering
\begin{tabular}{ c|c|c|c|c } 
 \hline
  Model & natural & PGD$_{\epsilon=2}$ & PGD$_{\epsilon=5}$ & PGD$_{\epsilon=8}$ \\
  %&  & (\epsilon = 1) & (\epsilon = 2.24)\\
 \hline
 \hline
Vanilla trained & 93.7\% & 45.0\%& 15.9\% & 5.6\%\\ 
Trades($1/\lambda=0.05$) & 91.4\%  & 53.5\% & 25.8\% & 8.6\% \\
Trades($1/\lambda=0.01$) & 92.6\% & 49.4\% & 19.2\% & 6.3\% \\
%Madry's & 83.53\% & 82.44\% & 76.22\% & 56.61\% \\
Ours & 93.8\% & 64.9\%  & 31.0\% & 10.9\% \\
 \hline
\end{tabular}
\label{Trades}
%\end{adjustbox}
\end{table}

\begin{table}[t]%\smaller
\caption{Accuracy-Robustness performance against PGD attacks on CIFAR-10, with various backbone models.}
\begin{center}
\begin{tabular}{ c||c||c|c|c} 
 \hline
 Models & natural & PGD$_{\epsilon=2}$ & PGD$_{\epsilon=5}$ & PGD$_{\epsilon=8}$\\
 % & Dataset & $(\epsilon = 2)$ & $(\epsilon = 5)$ & $(\epsilon = 8)$ \\ 
 \hline
 \hline
 ResNet-18 (vanilla)& 93.1\%&44.1\%& 14.6\% & 5.8\%\\ 
 & (-0.7\%)& (+16.3\%)& (+10.9\%) & (+2.7\%)\\ 
 ResNet-18 (ours) & 92.4\% & 60.4\%& 25.5\% & 8.5\%  \\ 
  \hline

 ResNet-34 (vanilla) & 93.3\%& 46.6\%& 15.3\% & 4.5\%\\
  & (-0.4\%) & (+14.8\%)& (+12.6\%) & (+4.4\%)\\ 
 ResNet-34 (ours) & 92.9\% & 61.4\%& 27.9\% & 8.9\%\\  
  \hline

 ResNet-50 (vanilla) & 93.7\% & 45.0\%& 15.9\% & 5.6\%\\ 
  & (+0.1\%) & (+19.9\%)& (+15.1\%) & (+5.3\%)\\ 
 ResNet-50 (ours) & 93.8\% & 64.9\%& 31.0\%& +10.9\%\\ 
  \hline
 ResNet-101 (vanilla) & 93.7\% & 46.9\%& 15.9\% & 6.8\%\\ 
  & (+0.1\%) & (+22.9\%) & (+16.3\%) & (+7.4\%)\\ 
 ResNet-101 (ours) & 93.8\% & 69.8\%& 32.3\%& +14.2\%\\ 
  \hline
\end{tabular}
\label{finaltable}
\end{center}
\end{table}
For the more complex dataset, CIFAR-10, Table~\ref{Trades} shows the results with ResNet-50 as the backbone. Unlike the MNIST dataset, ResNet-50 with the proposed adversarial training strategy significantly outperforms TRADES in terms of adversarial robustness. Note that in our experiment, the maximal $\epsilon$ of DRA in adversarial training is 2, which corresponds to $\epsilon=4-5$ in PDG attacks. We believe that training the model with a large DRA $\epsilon$ would further improve the robustness.

In addition, we vary the backbone models for the CIFAR-10 experiments and report the accuracy-robustness performance in Table~\ref{finaltable}. First, for clean CIFAR-10 image classification, vanilla-trained models and our adversarial fine-tuned models achieve comparable performance in all examined backbone models. %In other words, our adversarial training strategy does not necessarily hurt the model accuracy compare to traditional adversarial training. 
In particular, ResNet-50 and ResNet-101 can even outperform their vanilla training counterparts in terms of clean data accuracy. Second, to defend against PGD attacks, we observe that models with larger capacity are able to boost their adversarial robustness to a larger margin, which supports our hypothesis discussed in section 3. %To preserve the standard performance, our method facilitate the boost of adversarial robustness over TRADES. Note that

\subsection{Effect of DRA on Adversarial Training}
Generation of strong adversarial samples (with higher attack success rates) is a critical factor in adversarial training. In this experiment, we investigate the impact of different types of adversarial samples in three different adversarial training strategies: Madry's~\cite{Madry}, TRADES~\cite{thero}, and ours. Concretely, we replace PGD samples with DRA data in Madry's, and TRADES. Similarly, we use PDG in the proposed method and compare its performance with DRA. We note that DRA is a stronger adversary than PGD for the same $\epsilon$ conditions due to the presence of soft-bounded constraint. To make a fair comparison, we choose different $\epsilon$ for them so that DRA-generated samples and PGD-generated samples can have similar attack strengths. Specifically, we set $\epsilon = 1$ for DRA and $\epsilon = 2.55$ for PGD, as both settings resulted in a robust accuracy of 44.5\% on vanilla-trained ResNet-50.%Since DRA is an independent adversarial generation method, we compare the proposed DRA algorithm with PGD, the most widely used adversarial attack method in literature, and investigate their impact on various adversarial training methods.

Table.~\ref{DRAvsPGD2} reports CIFAR-10 classification performance with ResNet-50 as the backbone model. We observe that in all three settings, DRA is more beneficial for improving model robustness. Furthermore, the clean data accuracy of DRA trained models is higher than that of the PGD trained models using Madry's and our method. Since the loss function of TRADES (Eqn.\ref{TRADES}) is essentially designed to improve robustness through trading accuracy, we believe it is reasonable to assume that the standard accuracy of the DRA trained model is slightly lower than that of the model trained with PGD. In short, Table.~\ref{DRAvsPGD2} with numerical results validate our hypothesis that DRA is a better adversary that benefits model's adversarial training.

% \begin{table}[htbp]\smaller
% \caption{CIFAR-10 Classification accuracy $(\%)$ by ResNet-18. Compared to PGD, DRA is a better candidate for adversarial training. We highlight the best model in each column.}
% \begin{center}
% \begin{tabular}{ c|c|c|c|c } 
%  \hline
%   Model & Clean data & PGD ($\epsilon=2$) & PGD($\epsilon=5$) & PGD($\epsilon=8$) \\
%   %&  & (\epsilon = 1) & (\epsilon = 2.24)\\
%  \hline
%  \hline
%  Vanilla & \textbf{93.07\%} & 39.80\% & 39.73\% \\ 
%  & $(-1.91\%)$ & $ (+10.36\%)$ & $(+14.13\%)$ \\
% PGD + fine-tune & 91.16\%  & 50.16\% & 53.86\% \\
%  & $(-0.65\%)$ & $ (+17.43\%)$ & $(+16.70\%)$ \\
% DRA + fine-tune & 92.42\% & \textbf{57.23\%} & \textbf{56.43\%}  \\ 
%  \hline
% \end{tabular}
% \label{DRAvsPGD2}
% \end{center}
% \end{table}
\begin{table}[t]%\smaller
\caption{ResNet-50 trained with PGD Vs. DRA on CIFAR-10}
\begin{center}
\begin{tabular}{ c|c|c|c|c } 
 \hline
  Model & Clean data & PGD$_{\epsilon=2}$ & PGD$_{\epsilon=5}$ & PGD$_{\epsilon=8}$ \\
  %&  & (\epsilon = 1) & (\epsilon = 2.24)\\
 \hline
 \hline
 Madry's + PGD & 88.8\% & 66.1\% & 30.9\% & 11.0\% \\ 
    &(+0.6\%) &(+0.6\%) &(+0.7\%) &(+1.5\%)\\
 Madry's + DRA& 89.4\% & 66.7\% & 31.6\% & 12.5\% \\ 
 \hline
TRADES + PGD & 87.9\%  & 65.0\% & 34.3\% & 12.1\%\\
    &(-0.7\%) &(+3.3\%) &(+4.8\%) &(+1.8\%)\\
TRADES + DRA & 87.2\%  & 68.3\% & 39.1\% & 13.9\%\\
\hline
ours + PGD & 90.6\% & 59.2\% & 27.0\% & 10.1\%\\ 
    &(+0.7\%) &(+0.8\%) &(+1.9\%) &(-0.3\%)\\
ours + DRA & 91.3\% & 60.0\% & 28.9\% & 9.8\%\\ 
 \hline
\end{tabular}
\label{DRAvsPGD2}
\end{center}
\end{table}

\subsection{Ablation on DRA Hyperparameter Setting}
Our DRA attack algorithm filters out unimportant pixels by a pre-fixed percentage and applies adversarial noises only to important image features that contribute to the final prediction. In this ablation study, we investigate the effect of hyperparameter settings (e.g. the value of significant feature percentage $p$ and noise budget $\epsilon$ in Algorithm~\ref{DRA}) on CIFAR-10 dataset. Specifically, CIFAR-10 images are in size of $3 \times 32 \times 32$. We treat the values of red, green and blue independently and thus obtain $N=3072$ pixel values per image. To comprehensively study this problem, we vary the values of $p$ and $\epsilon$ and report DRA's attack success rates on CIFAR-10 vanilla-trained ResNet-18, ResNet-34 and ResNet-50.

\begin{figure}[t]%\small
 % \centering
  \subfloat[$\epsilon$ = 1]{\includegraphics[width=0.24\textwidth]{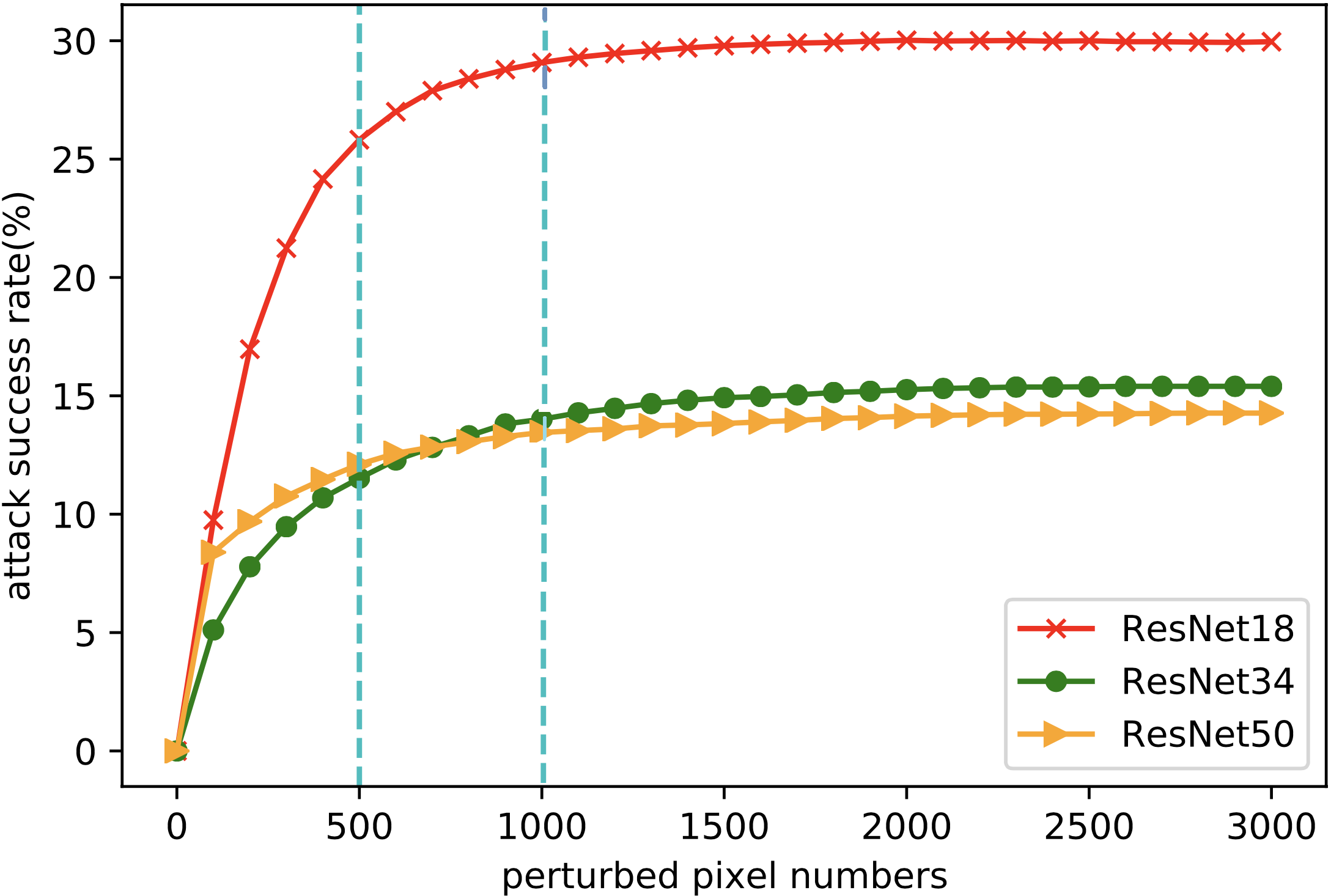}}%\label{T1}}  
  \subfloat[$\epsilon$ = 2] {\includegraphics[width=0.24\textwidth]{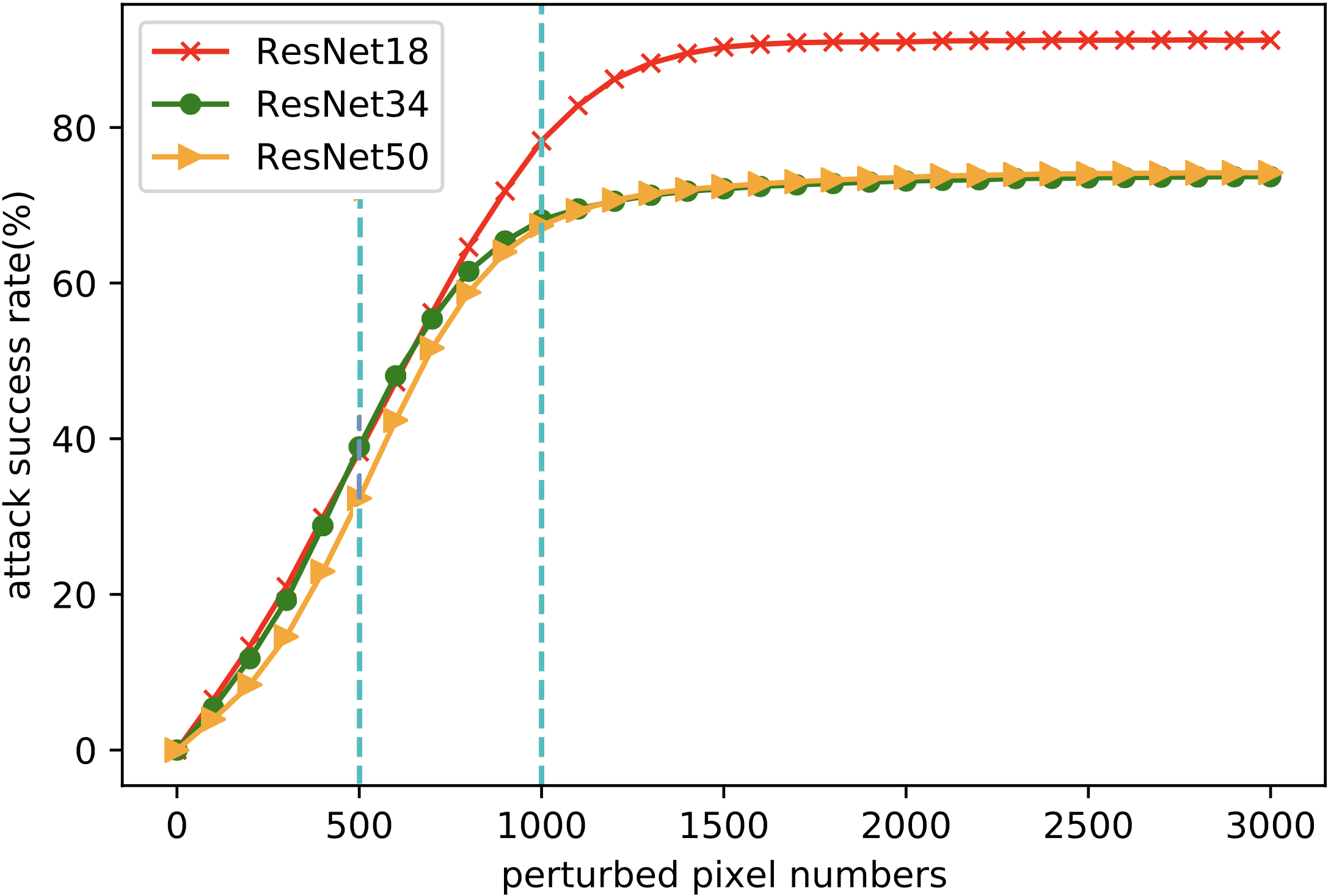}}%\label{T2}}
  \caption{Ablation on DRA hyperparameter settings: DRA's attack success rates versus the significant feature percentage $p$ on CIFAR-10 vanilla-trained ResNet models. A marginal improvement on attack success rate is observed when $p>1/3$.}
  \label{ablation}
\end{figure}

Fig.~\ref{ablation} reports adversarial attack success rate versus the percentage $p$ out of the 3072 pixels. With the settings $\epsilon=1$, we note that the successful attack rate of DRA grows linearly in the most significant 500 (i.e. $p=1/6$) pixels and saturates in about 1000 ($p=1/3$) pixels. For $\epsilon=2$, the successful attack rate also almost saturates at approximately 1000 ($p=1/3$) pixels. %

%Therefore, to avoid excessive learning load that may cause degradation to accuracy, choosing the value $p$ according to noise budget and capacity of the model is important. For example, when training ResNet-18 with $\epsilon=2$, we can observe that $p=1/3$ is an optimal choice for balancing the learning complexity. 
To further investigate the impact of DRA thresholds on the overall adversarial training method, we train ResNet-50 with different values of $p$ and report the performance in Table.~\ref{thres}. We notice that a higher threshold value $p$ does contribute to better model robustness, however, the improvement becomes marginal as $p>1/3$. In particular, increasing $p$ from 1/2 to 1 only leads to less than 0.5\% robustness increment for PGD attacks, but causes the downgrade of natural accuracy by 0.8\%. These results also support our claim that introducing too much unnecessary noise in the images complicates model optimization and tends to lead to a standard performance loss. Similar to the results in Fig. \ref{ablation}, Table \ref{thres} indicates that $p=1/3-1/2$ is a good setting for generating adversarial samples in the proposed method. 

\begin{table}[t]
\caption{Ablation on DRA hyperparameter settings: adversarial training performance versus different threshold value $p$.}
%\begin{adjustbox}{width=0.8\columnwidth,center}
\centering
\begin{tabular}{ c|c|c|c|c } 
 \hline
  $p$ & natural & PGD$_{\epsilon=2}$ & PGD$_{\epsilon=5}$ & PGD$_{\epsilon=8}$ \\
  %&  & (\epsilon = 1) & (\epsilon = 2.24)\\
 \hline
 \hline
0 & 93.7\% & 45.0\% & 15.9\% & 5.6\% \\ 
1/6 & 91.8\% & 58.5\% & 28.0\% & 9.2\% \\ 
1/3 & 91.3\%  & 60.0\% & 28.9\% & 9.8\% \\
1/2 & 90.4\% & 61.4\% & 29.1\% & 10.0\% \\
1 & 89.6\% & 61.9\% & 29.3\% & 10.1\% \\
 \hline
\end{tabular}
\label{thres}
%\end{adjustbox}
\end{table}

% In addition, we have an interesting observation from Fig.~\ref{ablation}. Recent works show that high capacity models are more robust than simple models \cite{b2,Madry}. For instance, ResNet-34 is more robust than ResNet-18 in Fig.~\ref{ablation}. However, despite the $16$-layer difference in architecture, ResNet-50 and ResNet-34 exhibit similar adversarial robustness. We notice that ResNet-50 has the same amount of residue blocks as ResNet-34, but it uses a 3-layer bottleneck architecture instead of a 2-layer basic residual block. If we treat a residue block as one processing unit, ResNet-50 is no deeper than ResNet-34. We hypothesize that this is the reason for the tie performance between ResNet-34 and ResNet-50. 

\subsection{Evaluation on Naturally Corrupted Images}

%\captionsetup[figure]%{font=small}
\begin{figure}[t]
\centerline{\includegraphics[width=\columnwidth]{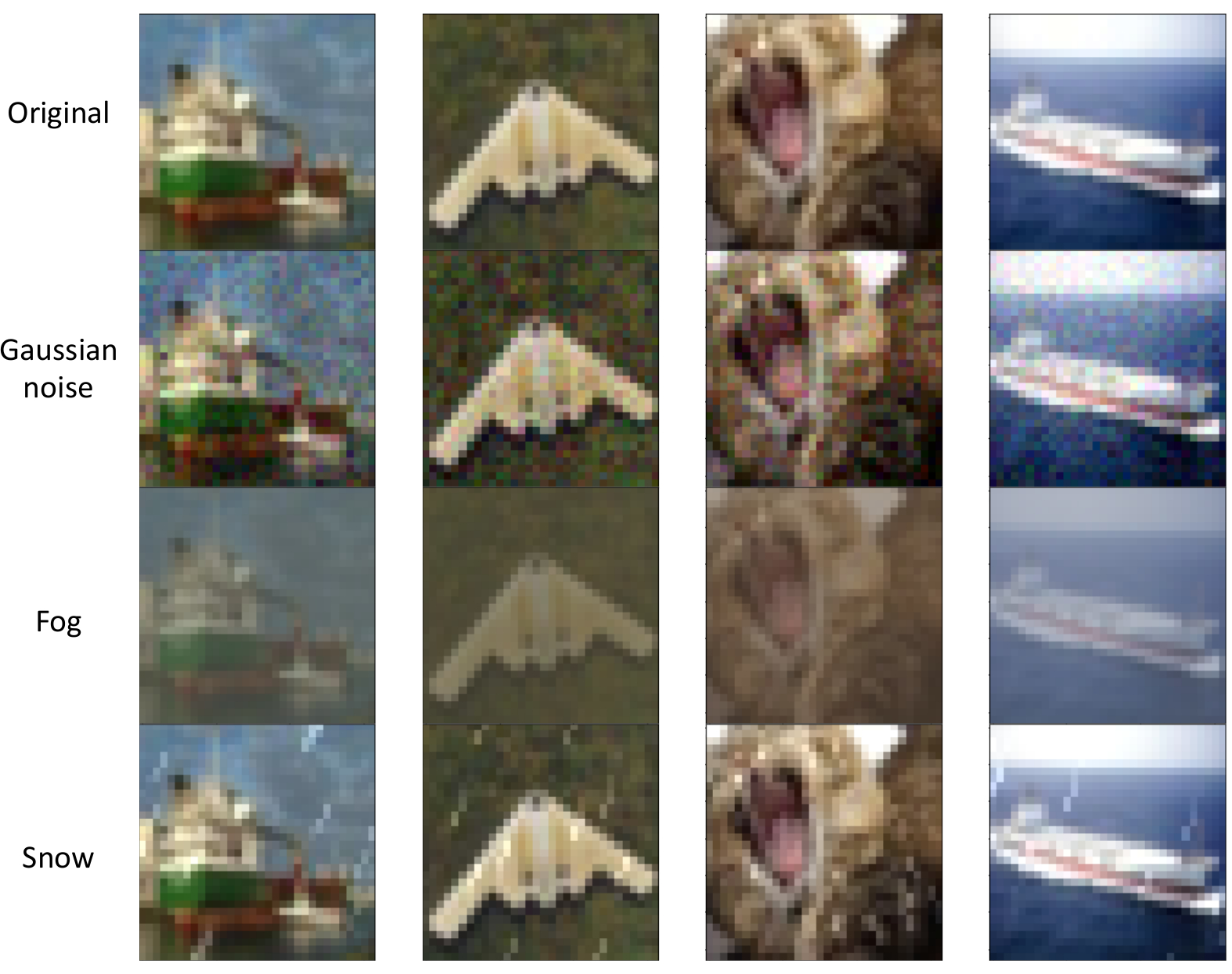}}
\caption{Corrupted image samples in\textit{ Cifar-10-C}~\cite{Cifar-10-c}.}
\label{corr}
\end{figure}

\begin{table}[t]%\smaller
\caption{The accuracy of DRA trained ResNet-50 vs. Vanilla trained ResNet-50 over different corrupted types.}
\begin{center}
\begin{tabular}{ c|c|c } 
 \hline
  Corrupted Type& ResNet-50  & ResNet-50   \\
  & (DRA Trained)&(Vanilla trained) \\
 \hline
 \hline
 Snow & 85.37\% & 84.10\%  \\ 
 Frost & 84.4\% & 81.41\%  \\ 
 Zoom$\_$blur & 84.97\% & 82.13\%  \\ 
 Motion$\_$blur & 76.89\% & 74.08\%  \\ 
 JPEG compression & 89.51\% & 84.51\% \\
 Gaussian noise & 75.85\% & 50.92\% \\
 \hline
\end{tabular}
\label{corrupt}
\end{center}
\end{table}

Corrupted data with naturally occurring perturbations and distributed shifts pose challenges to model generalization. In this experiment, we explore the potential of DRA fine-tuned models on corrupted data. In this experiment, we train a ResNet-50 model with our adversarial fine-tuning strategy and evaluate its performance on the corrupted CIFAR-10 dataset (\textit{CIFAR-10-C}~\cite{Cifar-10-c}). In CIFAR-10-C, the clean CIFAR-10 data are processed to mimic various image distortions under harsh conditions. Table~\ref{corrupt} presents classification accuracy on the CIFAR-10-C dataset with the DRA refined model and the vanilla trained model. Results show that the DRA fine-tuned model exhibits stronger robustness to different types of corruption.

\section{Conclusion}
In this paper, we aim to tackle a unique problem in adversarial training: improving the adversarial robustness of a model without sacrificing the standard performance. We explicitly formulate the problem and propose a cost-efficient adversarial training strategy. It decomposes adversarial training into two phases: standard training and robust training. In addition, we introduce a training-friendly adversary to further benefit adversarial training. Extensive experimentation on MNIST and CIFAR-10 datasets suggests that the proposed adversarial training strategy serves better for the target objective. %This allows us to improve the robustness of the model without disturbing its generalizability of clean data. While traditional adversarial training focuses on achieving high robustness, our objective is to use adversarial samples as a fair regulation method that helps models benefit in terms of both accuracy and robustness at the same time. 
%Furthermore, we provide solid theoretical and empirical evidence that model capacity is the main constraint that prevents adversarial training from learning a classifier with both optimal standard and robust accuracy. Therefore, the flexibility offered by DRA makes it an ideal candidate to meet our objective.

\printbibliography

@inproceedings{mart,
  title={IMPROVING ADVERSARIAL ROBUSTNESS REQUIRES
REVISITING MISCLASSIFIED EXAMPLES},
  author={Yisen Wang1 and Difan Zou2 and Jinfeng Yi3 and James Bailey4 and Xingjun Ma4 and Quanquan Gu},
  booktitle={International Conference on Learning RepresentationsInternational Conference on Learning Representations},
  year={2020}
}

@inproceedings{b1,
  title={Deep neural networks are easily fooled: High confidence predictions for unrecognizable images},
  author={Nguyen, Anh and Yosinski, Jason and Clune, Jeff},
  booktitle={Conference on Computer Vision and Pattern Recognition},
  pages={427--436},
  year={2015}
}

@inproceedings{b2,
  title={Intriguing properties of neural networks},
  author={Szegedy, Christian and Zaremba, Wojciech and Sutskever, Ilya and Bruna, Joan and Erhan, Dumitru and Goodfellow, Ian and Fergus, Rob},
  booktitle={International Conference on Learning Representations},
  year={2014}
}

@inproceedings{b3,
  title={Explaining and harnessing adversarial examples},
  author={Goodfellow, Ian J and Shlens, Jonathon and Szegedy, Christian},
  journal={International Conference on Learning Representations},
  year={2015}
}

@ainproceedings{b4,
  title={Adversarial machine learning at scale},
  author={Kurakin, Alexey and Goodfellow, Ian and Bengio, Samy},
  journal={International Conference on Learning Representations},
  year={2017}
}

@inproceedings{b5,
  title={Ensemble adversarial training: Attacks and defenses},
  author={Tram{\`e}r, Florian and Kurakin, Alexey and Papernot, Nicolas and Goodfellow, Ian and Boneh, Dan and McDaniel, Patrick},
  journal={International Conference on Learning Representations},
  year={2018}
}

@inproceedings{b6,
  title={Boosting adversarial attacks with momentum},
  author={Dong, Yinpeng and Liao, Fangzhou and Pang, Tianyu and Su, Hang and Zhu, Jun and Hu, Xiaolin and Li, Jianguo},
  booktitle={Conference on Computer Vision and Pattern Recognition},
  pages={9185--9193},
  year={2018}
}

@article{simple,
  title={A Simple Fine-tuning Is All You Need: Towards Robust Deep Learning Via Adversarial Fine-tuning},
  author={Jeddi, Ahmadreza and Shafiee, Mohammad Javad and Wong, Alexander},
  journal={arXiv preprint arXiv:2012.13628},
  year={2020}
}

@inproceedings{ft1,
  title={Adversarial robustness: From self-supervised pre-training to fine-tuning},
  author={Chen, Tianlong and Liu, Sijia and Chang, Shiyu and Cheng, Yu and Amini, Lisa and Wang, Zhangyang},
  booktitle={Proceedings of the IEEE/CVF Conference on Computer Vision and Pattern Recognition},
  pages={699--708},
  year={2020}
}

@inproceedings{ft2,
  title={Using pre-training can improve model robustness and uncertainty},
  author={Hendrycks, Dan and Lee, Kimin and Mazeika, Mantas},
  booktitle={International Conference on Machine Learning},
  pages={2712--2721},
  year={2019},
  organization={PMLR}
}

@inproceedings{Madry,
  title={Towards deep learning models resistant to adversarial attacks},
  author={Madry, Aleksander and Makelov, Aleksandar and Schmidt, Ludwig and Tsipras, Dimitris and Vladu, Adrian},
  booktitle={arXiv preprint arXiv:1706.06083},
  year={2017}
}

@inproceedings{hurt,
  title={Adversarial training can hurt generalization},
  author={Raghunathan, Aditi and Xie, Sang Michael and Yang, Fanny and Duchi, John C and Liang, Percy},
  booktitle={arXiv preprint arXiv:1906.06032},
  year={2019}
}

@inproceedings{odds,
  title={Robustness may be at odds with accuracy},
  author={Tsipras, Dimitris and Santurkar, Shibani and Engstrom, Logan and Turner, Alexander and Madry, Aleksander},
  booktitle={arXiv preprint arXiv:1805.12152},
  year={2018}
}

@inproceedings{closerlook,
  title={A closer look at accuracy vs. robustness},
  author={Yang, Yao-Yuan and Rashtchian, Cyrus and Zhang, Hongyang and Salakhutdinov, Ruslan and Chaudhuri, Kamalika},
  booktitle={Conference on Neural Information Processing Systems},
  volume={33},  
  year={2020}
}

@inproceedings{thero,
  title={Theoretically principled trade-off between robustness and accuracy},
  author={Zhang, Hongyang and Yu, Yaodong and Jiao, Jiantao and Xing, Eric and El Ghaoui, Laurent and Jordan, Michael},
  booktitle={International Conference on Interactive Mobile Communication, Technologies and Learning},
  pages={7472--7482},
  year={2019},
  organization={Proceedings of Machine Learning Research}
}

@article{bugs,
  title={Adversarial examples are not bugs, they are features},
  author={Ilyas, Andrew and Santurkar, Shibani and Tsipras, Dimitris and Engstrom, Logan and Tran, Brandon and Madry, Aleksander},
  journal={arXiv preprint arXiv:1905.02175},
  year={2019}
}

@misc{b11,
  title={Adversarial examples in the physical world},
  author={Kurakin, Alexey and Goodfellow, Ian and Bengio, Samy and others},
  booktitle={International Conference on Learning Representations},
  year={2017}
}

@inproceedings{b13,
  title={Delving into transferable adversarial examples and black-box attacks},
  author={Liu, Yanpei and Chen, Xinyun and Liu, Chang and Song, Dawn},
  booktitle={International Conference on Learning Representations},
  year={2017}
}

@inproceedings{Xie,
  title={Adversarial examples improve image recognition},
  author={Xie, Cihang and Tan, Mingxing and Gong, Boqing and Wang, Jiang and Yuille, Alan L and Le, Quoc V},
  booktitle={Conference on Computer Vision and Pattern Recognition},
  pages={819--828},
  year={2020}
}

@inproceedings{downgrade1,
  title={Unlabeled data improves adversarial robustness},
  author={Carmon, Yair and Raghunathan, Aditi and Schmidt, Ludwig and Liang, Percy and Duchi, John C},
  booktitle={Conference on Neural Information Processing Systems},
  year={2019}
}

@inproceedings{downgrade2,
  title={Cat: Customized adversarial training for improved robustness},
  author={Cheng, Minhao and Lei, Qi and Chen, Pin-Yu and Dhillon, Inderjit and Hsieh, Cho-Jui},
  booktitle={arXiv preprint arXiv:2002.06789},
  year={2020}
}

@inproceedings{downgrade3,
  title={Instance adaptive adversarial training: Improved accuracy tradeoffs in neural nets},
  author={Balaji, Yogesh and Goldstein, Tom and Hoffman, Judy},
  booktitle={arXiv preprint arXiv:1910.08051},
  year={2019}
}

@inproceedings{fgm1,
  title={Learning with a strong adversary},
  author={Huang, Ruitong and Xu, Bing and Schuurmans, Dale and Szepesv{\'a}ri, Csaba},
  booktitle={arXiv preprint arXiv:1511.03034},
  year={2015}
}

@article{overfit,
  title={Adversarially robust generalization requires more data},
  author={Schmidt, Ludwig and Santurkar, Shibani and Tsipras, Dimitris and Talwar, Kunal and Madry, Aleksander},
  journal={Advances in neural information processing systems},
  volume={31},
  year={2018}
}

@inproceedings{cifar-10-c,
  title={Benchmarking neural network robustness to common corruptions and perturbations},
  author={Hendrycks, Dan and Dietterich, Thomas},
  journal={International Conference on Learning Representations},
  year={2019}
}

@inproceedings{inDistributionRobustness,
  title={Adversarial vulnerability for any classifier},
  author={Fawzi, Alhussein and Fawzi, Hamza and Fawzi, Omar},
  booktitle={Conference on Neural Information Processing Systems},
  pages={1186–1195},
  year={2018},
  organization={Conference on Neural Information Processing Systems}
}

@article{selfie,
  title={Selfie: Self-supervised pretraining for image embedding},
  author={Trinh, Trieu H and Luong, Minh-Thang and Le, Quoc V},
  journal={arXiv preprint arXiv:1906.02940},
  year={2019}
}

@inproceedings{jigsaw,
  title={Unsupervised learning of visual representations by solving jigsaw puzzles},
  author={Noroozi, Mehdi and Favaro, Paolo},
  booktitle={European conference on computer vision},
  pages={69--84},
  year={2016},
  organization={Springer}
}

@inproceedings{Raghunathan,
  title={Understanding and Mitigating the Tradeoff Between Robustness and Accuracy},
  author={Raghunathan, Aditi and Xie, Sang Michael and Yang, Fanny and Duchi, John and Liang, Percy},
  booktitle={Proceedings of Machine Learning Research},
  pages={7909-7919},
  year={2020},
  organization={Proceedings of Machine Learning Research}
}

@inproceedings{Stutz,
  title={Disentangling Adversarial Robustness and Generalization},
  author={Stutz, David and Hein, Matthias and Schiele, Bernt},
  booktitle={Conference on Computer Vision and Pattern Recognition},
  pages={6976-6987},
  year={2019},
  organization={Conference on Computer Vision and Pattern Recognition}
}

@inproceedings{PAMI,
  title={Universal Adversarial Attack on Attention and the Resulting Dataset DAmageNet},
  author={Chen, Sizhe and He, Zhengbao and Sun, Chengjin and Yang, Jie and  Huang, Xiaolin},
  booktitle={IEEE Trans. Pattern Analysis and Machine Intelligence},
  year={2020}
}

\end{document}